\documentclass{article}
\usepackage{silence}
\usepackage[preprint]{neurips_2026}
\usepackage[preprint]{neurips_2026}
\usepackage{graphicx}
\usepackage[utf8]{inputenc} 
\usepackage[T1]{fontenc}    
\usepackage{hyperref}       
\usepackage{url}            
\usepackage{booktabs}       
\usepackage{amsfonts}       
\usepackage{nicefrac}       
\usepackage{microtype}      
\usepackage{xcolor}         
\usepackage{caption, subcaption}
\usepackage{booktabs}
\usepackage{float}
\usepackage{multirow}
\usepackage[table]{xcolor}
\usepackage{arydshln}
\usepackage{paralist}
\usepackage{amsmath}
\usepackage{tabularray}
\usepackage{algorithm}
\usepackage{algpseudocode}
\newtheorem{definition}{Definition}

\title{Query-aligned video frame selection\\ for long video understanding}

\author{%
Md. Safayet Islam\\
  Department of Computer Science\\
  University of Miami\\
  Coral Gables, FL 33146 \\
  \texttt{mxi451@miami.edu}
  \And
  Dilip Sarkar\\
  Department of Computer Science\\
  University of Miami\\
  Coral Gables, FL 33146 \\
  \texttt{sarkar@miami.edu} \\
  \And
  Liang Liang\\
  Department of Computer Science\\
  University of Miami\\
  Coral Gables, FL 33146 \\
  \texttt{liang.liang@miami.edu} \\
}
\begin{document}
\maketitle
\begin{abstract}
Multimodal large language models (MLLMs) process multimodal inputs by converting text, images, and videos into token sequences that are subsequently processed by a backbone language model. However, the limited visual token budget of these models makes long-form video understanding challenging. While MLLMs have achieved excellent performance in understanding the content of individual images, video understanding remains significantly more difficult because of the large amount of visual information contained in videos. For instance, a 5-minute video at 24 frames per second (fps) contains 7,200 frames. In practice, however, MLLMs typically process only a small subset of these frames, usually ranging from 8 to 64. These frames are commonly sampled uniformly, regardless of their relevance to the question being answered. To address this limitation, several training-free, model-agnostic methods for selecting question-relevant frames have recently been proposed to improve video question answering.

In this work, we introduce a frame-selection method designed specifically for multiple-choice questions. We extend the query text by appending semantic cues derived from the answer choices and employ a direct query-frame alignment scoring mechanism. To the best of our knowledge, our method is the first to directly utilize answer choices as inference-time cues for selecting frames relevant to answering a question. The method first constructs a compact candidate pool by subsampling video frames at a fixed rate. The frames are then scored according to their maximum cosine similarity across all question-answer pairs to identify the most relevant frames for a given query. This approach preserves a fixed token budget while improving the relevance of the visual evidence provided to the downstream MLLM.

We evaluate the effectiveness of our frame-selection method on the MLVU, Video-MME, and LongVideoBench benchmarks using three MLLMs: LLaVA-Mini, Qwen2-VL, and LLaVA-Video. Experimental results demonstrate that answer-aware frame selection generally outperforms uniform sampling and existing training-free frame-selection methods under the same frame budget.
\end{abstract}

\section{Introduction}
\label{sec:intro}

\begin{figure}[htb]
     \centering
     \includegraphics[width=0.95\linewidth]{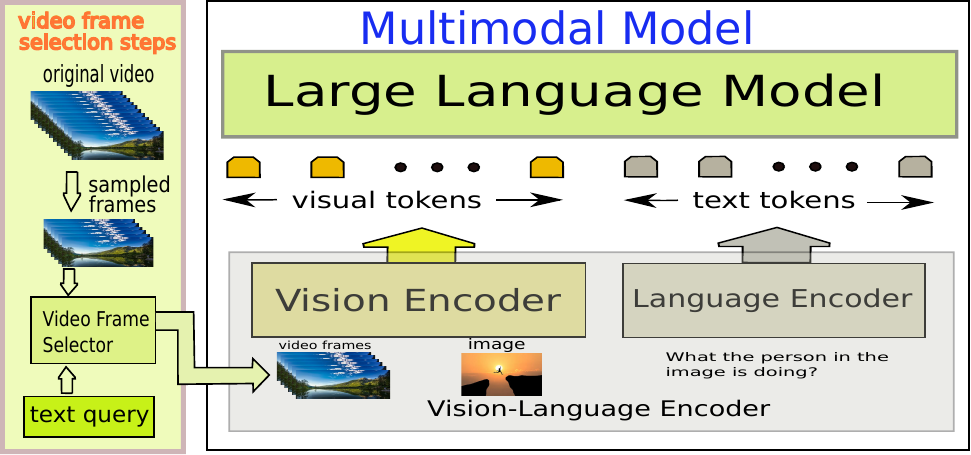}
     \caption{Block diagram of a video frame selector for a multimodal large language model (MLLM). The left column illustrates the three main steps of the video frame-selection process. The proposed frame-selection module is training-free, MLLM-agnostic, and plug-and-play.}
     \label{fig:MLLMandFrameSelector}
 \end{figure}

Multimodal Large Language Models (MLLMs), also known as multimodal models, extend Large Language Models (LLMs) to understand both visual and textual inputs. An MLLM typically consists of a front-end encoder and fusion modules~\cite{lin2023_video_llava,Qwen2VLwang2024qwen2_vl,LLaVAMiniZhang2025llava} that convert input text, images, and videos into a sequence of tokens (see Fig.~\ref{fig:MLLMandFrameSelector}). For example, VideoLLaMA2~\cite{VideoLLaMA2Cheng2024videollama2} can process approximately 2,000 tokens, whereas VILA-V1.5~\cite{lin2024vila} supports approximately 4,000 tokens.
 \par
The input-size constraint is less restrictive for images, for which these MLLMs have achieved excellent performance in understanding image content. However, this constraint poses a significant challenge for long-video understanding. A 5-minute video at 24 frames per second (fps) contains 7,200 frames, far exceeding the number of frames that can be processed during practical MLLM inference. For long-video understanding, the most common approach is to sample frames uniformly at regular intervals. However, this approach may overlook frames containing critical visual cues required to answer a question correctly, as it does not account for the specific question being asked.

 \begin{figure}[htb]
    \centering

    \includegraphics[width=\textwidth]{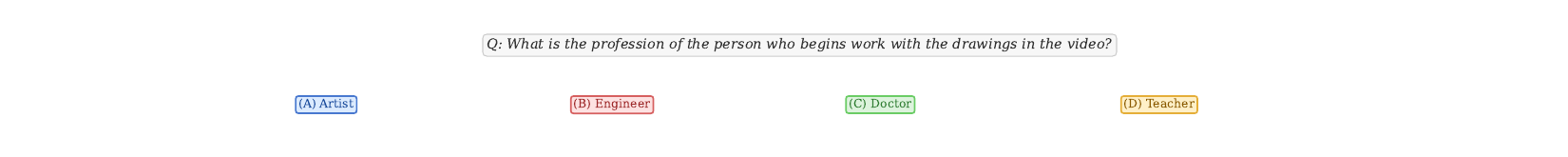}

    \vspace{-0.1em}

    \begin{subfigure}[t]{0.48\textwidth}
        \centering
        \includegraphics[width=\linewidth]{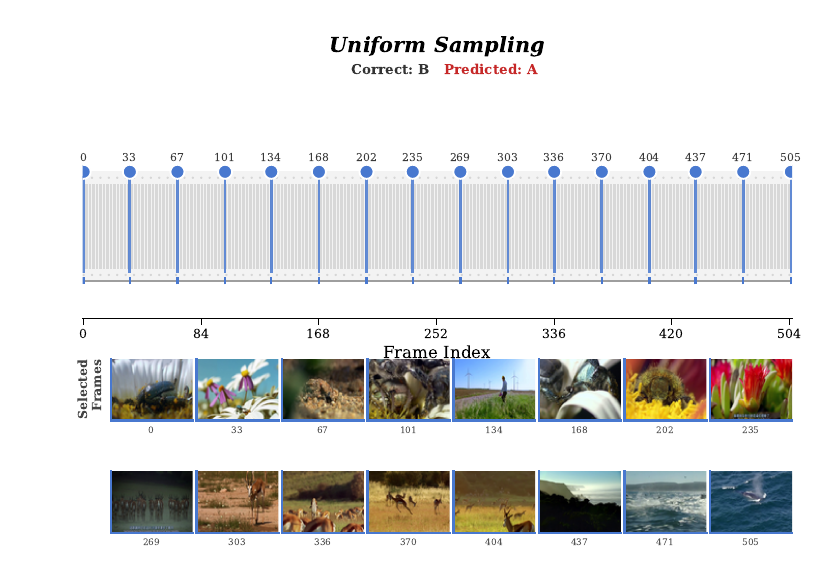}
        \caption{Uniform interval-based frame selection.}
        \label{fig:uniformSampling}
    \end{subfigure}
    \hfill
    \begin{subfigure}[t]{0.50\textwidth}
        \centering
        \includegraphics[width=\linewidth]{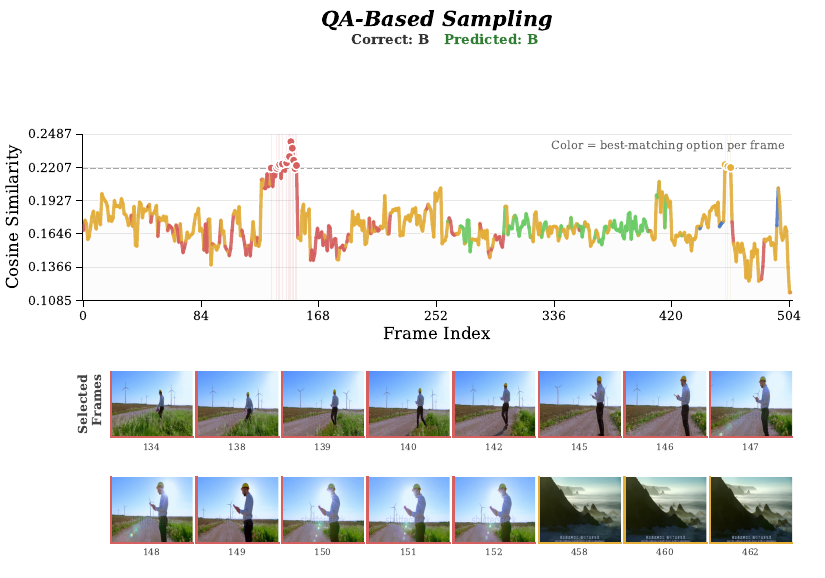}
        \caption{Question-answer-aligned frame selection.}
        \label{fig:QAbased}
    \end{subfigure}

    \caption{Comparison of uniform-interval-based and question-answer-aligned frame-selection methods using the proposed approach. Note that the proposed method selected 13 of the 16 relevant frames, whereas uniform-interval-based sampling selected only one.}
    \label{fig:EffectOfVideoFrameSelection}
\end{figure}

Fortunately, the answers to most video-related questions are often contained within a small number of typically consecutive video frames~\cite{zhu2026focus}. Thus, long-video question answering can be divided into two subproblems: \begin{inparaenum}[$(i)$] \item selecting video frames that are relevant to the question and \item generating an answer using an MLLM based on the selected frames and the query.
  
 \end{inparaenum}
 \par
Selecting meaningful frames from a long video can be computationally challenging. Decoding and evaluating every frame requires substantial storage, CPU resources, and GPU memory, resulting in increased inference time, particularly for longer videos~\cite{zhu2026focus}. Therefore, a practical frame-selection method should first reduce the candidate frame pool through fixed-rate sampling and then select a small subset of frames most relevant to answering the given question.
 \par

A common solution to the candidate-pool size problem is query-aware video frame downsampling~\cite{lin2023_sphinx,video_chatgptMaaz2024,wang2022_internvideo,QframeZhang2025q}. This approach is motivated by the observation that blind uniform downsampling, which is used to satisfy an MLLM's token budget, can miss frames that are relevant to the question~\cite{tang2025adaptive,QframeZhang2025q,zhu2026focus}. Query-aware frame selection addresses this limitation by leveraging the question to identify frames that are more likely to contain the necessary visual cues. Unlike uniform sampling, which selects frames solely based on their temporal positions, query-aware selection evaluates candidate frames according to their relevance to the question. For example, the uniformly sampled frames shown in Fig.~\ref{fig:uniformSampling} contain only one question-relevant frame. In contrast, query-aligned frame-selection methods assign scores to video frames and select those that receive higher alignment scores. As shown in Fig.~\ref{fig:QAbased}, the 16 frames selected using one of our proposed query-aligned selection methods contain 13 relevant frames.

\par
The left column in Fig.~\ref{fig:MLLMandFrameSelector} presents a high-level overview of a query-aware, training-free, MLLM-agnostic, and plug-and-play frame-selection module for selecting question-relevant video frames (see Figs.~\ref{fig:QVframework} and~\ref{fig:QAVframework} for details of the proposed frame-selection method). The framework consists of two independent but complementary modules: one for video-frame subsampling to reduce the frame-selection search space and the other for query-aware video-frame selection. Several recent methods follow the general framework shown in Fig.~\ref{fig:MLLMandFrameSelector}, including Q-Frame (CVF2025)~\cite{QframeZhang2025q}, AKS (CVPR2025)~\cite{tang2025adaptive}, and FOCUS (ICLR2026)~\cite{zhu2026focus}. AKS and FOCUS use BLIP~\cite{li2023blip2} to compute frame-alignment scores for frame selection. Q-Frame~\cite{QframeZhang2025q} generates visual and query tokens using CLIP~\cite{radford2021clip}, and the dot product between these tokens is subsequently used to guide frame selection. In this work, we introduce a new video-frame-selection method, which is described in the following paragraph.
 
\par
Our proposed method performs regular-interval frame subsampling, similar to Q-Frame, but differs in two key respects. First, we replace the inner-product ranking used in Q-Frame with cosine similarity, which is invariant to embedding magnitude and empirically improves downstream accuracy across many of the evaluated settings. Second, and to the best of our knowledge, we are the first to utilize the answer choices provided in multiple-choice questions as inference-time cues for frame selection. For each candidate frame, we compute its cosine similarity with every question-answer pair and assign the frame the maximum score obtained across all pairs (see Section~\ref{sec:QueryAnswerAligned}). This answer-aware scoring mechanism directs frame selection toward visual evidence that is relevant to what the answer looks like, rather than solely to what the question asks. To validate the effectiveness of the proposed frame-selection method, we evaluate it on three open-source MLLMs (LLaVA-Mini-8B~\cite{LLaVAMiniZhang2025llava}, Qwen2-VL-7B~\cite{Qwen2VLwang2024qwen2_vl}, and LLaVA-Video-7B~\cite{zhang2024llava-video}) using three video-understanding benchmarks: Video-MME~\cite{fu2024video_mme}, LongVideoBench~\cite{LongVideoBenchWu2024}, and MLVU~\cite{MLLVUbenchmarkZhou2024_mlvu}.

 \par
Our main contributions are fourfold: \begin{inparaenum}[(1)] \item We propose a cosine-similarity-based query-video frame-scoring method that uses token vectors generated by pretrained vision-language encoders; \item We introduce a query-answer-video (QAV)-aligned frame-selection method that incorporates multiple-choice answer options as inference-time cues through a max-over-options scoring strategy; \item Our method is training-free, MLLM-agnostic, and plug-and-play, while preserving a fixed downstream MLLM frame budget and restoring the selected frames to their original temporal order; and \item We validate the proposed method on three open-source MLLMs and three long-video-understanding benchmarks, including comparisons across question categories and video-duration ranges against recent frame-selection baselines. \end{inparaenum}

\section{Preliminaries and Related Work}
\label{sec:relatedWork} 
\subsection{Preliminaries}
\label{prelim}
Let $[n]$ denote a set of $n$ integers. Throughout this paper, scalars are denoted by non-bold italic letters, vectors by bold lowercase letters, and matrices by bold uppercase letters. Let $\mathcal{V}^{(G)} = (\mathbf{V}^{(G)}_i \mid i \in [T])$ denote a video consisting of a sequence of $T$ uncompressed video frames (i.e., images). For $[T_s] \subseteq [T]$, let $\mathcal{V}^{(S)} = (\mathbf{V}^{(G)}_i \mid i \in [T_s])$ denote a subsequence of $T_s$ frames sampled from $\mathcal{V}^{(G)}$. Similarly, for $[k] \subseteq [T_s]$, let $\mathcal{V}^{(M)} = (\mathbf{V}^{(S)}_i \mid i \in [k])$ denote a subsequence of $k$ frames selected from $\mathcal{V}^{(S)}$. Let $Q$ denote a text query, possibly associated with $n_a$ answer choices, $A = \{a_i \mid i \in [n_a]\}$. Let $Q$ and each $a_i \in A$ be paired to form a set of $n_a$ text sequences, $QA = \{Qa_i \mid i \in [n_a]\}$. Let $\mathcal{M}_{\theta_m}(\cdot,\cdot)$ denote an MLLM that processes $Q$ (or $Qa_i$) and $\mathcal{V}^{(M)}$ and produces an output $\hat{A}$; that is, \[ \hat{A} = \mathcal{M}_{{\theta}_{m}}(Q,\mathcal{V}^{(M)}). \] Let $\mathcal{VL}_{\theta_{vl}}(\cdot)$ denote a pretrained vision-language model, such as CLIP~\cite{radford2021clip} or LongCLIP~\cite{zhang2024longclip}, that converts an input video frame $\mathbf{V}$ into a token vector $\mathbf{vt}$ and a text query $Q$ into a token vector $\mathbf{qt}$. Note that $\mathbf{vt}$ and $\mathbf{qt}$ have the same dimensionality.

\begin{definition}[Cosine similarity]
    \label{def:cosSim}
    Let $\mathbf{x}$ and $\mathbf{y}$ be two nonzero vectors with identical dimension. 
    \begin{equation}
        \operatorname{CosSim}(\mathbf{x},\mathbf{y}) = \frac{\mathbf{x}^{\top}\mathbf{y}}{\lVert\mathbf{x}\rVert_2 \lVert\mathbf{y}\rVert_2}
        \label{eq:cosSim}
    \end{equation}
The denominator performs standard $\ell_2$ normalization of both vectors before measuring their alignment.
\end{definition}

\subsection{Related Work}
\label{relatedWork}
Three lines of work aim to improve video understanding: \begin{inparaenum}[(1)] \item training stronger MLLMs with more data, more parameters, or both~\cite{LLaVAMiniZhang2025llava,Qwen2VLwang2024qwen2_vl,zhang2024llava-video}; \item training or adapting the MLLM front end while keeping the language-model back end frozen~\cite{wang2023vaquita}, or first training the front end and then fine-tuning the back end~\cite{li2024monkey}; and \item developing training-free, model-agnostic, and plug-and-play frame-selection methods~\cite{tang2025adaptive,QframeZhang2025q,zhu2026focus}. \end{inparaenum} Because our method belongs to the third category, we briefly discuss the three most recent methods and use them as baselines for comparison in our experiments.

\par
The Q-Frame method~\cite{QframeZhang2025q} consists of the following steps. First, it obtains $T_s$ candidate frames by uniformly sampling a video containing $T$ frames. Next, it uses the CLIP~\cite{radford2021clip} vision-language model to generate visual token vectors $\mathbf{vt}_i$ for $i \in [T_s]$ from the downsampled video frames and a text token vector $\mathbf{qt}$ from the query text. It then computes $T_s$ inner-product scores, $I_i = \mathbf{qt}\cdot\mathbf{vt}_i$, for $i \in [T_s]$, converts these scores into a probability distribution using the \textit{softmax} function, injects Gumbel noise into the logarithm of the resulting probability distribution, and finally selects the top $k$ frames based on the resulting scores.
\par
Both AKS~\cite{tang2025adaptive} and FOCUS~\cite{zhu2026focus} evaluate frame-query alignment using the BLIP~\cite{li2023blip2} vision-language model. The AKS method adaptively samples frames to identify the best subset of frames, whereas FOCUS~\cite{zhu2026focus} first divides a video into disjoint segments, performs a coarse-grained exploration to identify the most promising segments, and then conducts a fine-grained exploration within those segments to select the most informative video frames. This strategy substantially reduces the search space.
\par
The initial video segmentation used by FOCUS is motivated by an empirical analysis of videos from two benchmark datasets~\cite{fu2024video_mme,LongVideoBenchWu2024}, which shows that video frames exhibit a high degree of temporal autocorrelation. Consequently, coarse-grained sampling is often sufficient to identify the most promising video segments. In this work, we compare these three methods against the proposed approach.

\section{Proposed Methods}
\label{sec:ProposedMethod}
\paragraph{Problem Statement}
Given a video $\mathcal{V}^{(G)}$, a text question $Q$, and an MLLM $\mathcal{M}_{\theta_m}(\cdot,\cdot)$ that can process the question $Q$ and at most $k$ video frames from $\mathcal{V}^{(G)}$ to generate an answer $\hat{A}$, the task is to select a possibly non-consecutive sequence of $k$ frames, $\mathcal{V}^{(M)}$, from the video $\mathcal{V}^{(G)}$ such that the probability of producing the correct answer to the question $Q$ is maximized.

\paragraph{Need for Video Sub-Sampling} 
A practical challenge is that processing long videos can exceed the storage, memory, and computational limits of many systems, particularly GPU memory. Fortunately, recent empirical studies suggest that query-relevant frames tend to be temporally localized: frames within a short temporal window, approximately five seconds according to the analysis presented in FOCUS~\cite{zhu2026focus}, often exhibit similar levels of query relevance. Thus, when frames are sampled at sufficiently fine regular intervals, such as one frame per second, the resulting candidate set is likely to contain at least some frames from the query-relevant portion of the video. In this work, we assume that regular-interval candidate sampling includes at least one query-relevant frame. Throughout this paper, $\mathcal{V}^{(S)}$ denotes the subset of frames sampled from the original video $\mathcal{V}^{(G)}$ at regular intervals.

\paragraph{Best Frame Subset}
In this work, we treat an MLLM as a black box that produces an answer to a question $Q$ after processing the question $Q$ and $k$ video frames from $\mathcal{V}^{(S)}$, without providing a confidence score for the generated answer. However, selecting the optimal frame subset requires a mechanism for estimating the confidence of the answer.
Let $R_{\mathcal{M}_{\theta_m}}(Q,\mathcal{V}^{(S)})$ denote a function that estimates the confidence of the answer produced by the MLLM. Using this function, we can define an ideal frame-selection function, $FS^{(\mathrm{ideal})}(Q,\mathcal{V}^{(S)})$, that selects the best frame subset $\mathcal{V}^{(M)}$ from a given video-frame sequence $\mathcal{V}^{(S)}$. A formal definition is given below:
\begin{equation}
FS^{(ideal)}(Q,\mathcal{V}^{(S)}) = \underset{[k] \subseteq [T_s]} {\arg \max} R_{\mathcal{M}_{{\theta}_{m}}}(Q,\mathcal{V}^{(M)})
\label{eq:bestFrameSecetion}
\end{equation}
There are two issues with the best frame-selection function in Eq.~\ref{eq:bestFrameSecetion}: \begin{inparaenum}[$(i)$] \item the combinatorial search space is intractable, and \item the confidence-estimation function $R_{\mathcal{M}_{\theta_{m}}}(\cdot,\cdot)$ is not available. \end{inparaenum}
\par
While the number of frames in the sampled set is much smaller, the computational cost of a brute-force implementation of Eq.~\ref{eq:bestFrameSecetion} remains prohibitively high in practice. Moreover, no reliable method exists for estimating the probability of obtaining the correct answer from a selected set of frames. Before seeking a practical solution, we must determine the potential utility of an individual video frame for answering a given question. In other words, we need a method for measuring the alignment between a video frame and the question. 
Because vision-language models are trained on image-text pairs~\cite{wang2023vaquita,liang2024keyvideollm}, they provide a natural means of estimating frame-question alignment scores and have been used for this purpose in prior work~\cite{QframeZhang2025q,tang2025adaptive,zhu2026focus}.

\paragraph{Query and Video Frame Alignment Score} 
Given a set of sampled video frames ($\mathcal{V}^{(S)}$), selecting the most promising frames ($\mathcal{V}^{(M)}$) requires a method for assessing each frame's relevance to the question being answered~\cite{zhu2026focus}. Existing frame-selection methods commonly use a vision-language model either to produce a relevance score directly or to generate a pair of semantic representations, one from the video frame and the other from the query text, whose alignment is then measured. Both AKS~\cite{tang2025adaptive} and FOCUS~\cite{zhu2026focus} obtain frame-query relevance scores using the BLIP vision-language model. Q-Frame~\cite{QframeZhang2025q} uses the inner product of a video-frame vector and a text-query vector. However, the inner product can produce inconsistent scores because it depends on the magnitude of the vectors as well as their relative orientation. In contrast to these methods, we show that cosine similarity can identify relevant video frames more effectively.

\paragraph{Temporal Order Preservation} 
Video frames ordered by query-frame alignment scores may not preserve their original temporal order in the video. After selecting the most promising frames, we reorder them according to their original temporal sequence to preserve the order of events in the video. This step is important even when all selected frames originate from the same video clip.

\subsection{Query-Video Aligned Frame Selection}
\label{sec:QueryAligned}
To bypass the issues described above in selecting the best frame subset, a proxy function that does not depend on $\mathcal{M}_{\theta_m}(\cdot,\cdot)$ is needed. One possible solution, which has been used in prior work and is also adopted in this work, is to employ a vision-language encoder such as CLIP~\cite{radford2021clip} or LongCLIP~\cite{zhang2024longclip}.
\par
Vision-language encoders offer an additional computational advantage for identifying relevant frames because they evaluate the relevance of each video frame independently. Consequently, after evaluating all $T_s$ video frames individually, one can select the $k$ frames with the highest potential to contribute to a correct answer. Let us denote a vision-language encoder by $\mathcal{VL}_{\theta_{vl}}(x)$, where $x$ represents either a video frame $\mathbf{V}$ or a text question $Q$.
\begin{figure}[htb]
    \centering
    \includegraphics[width=.75\textwidth]{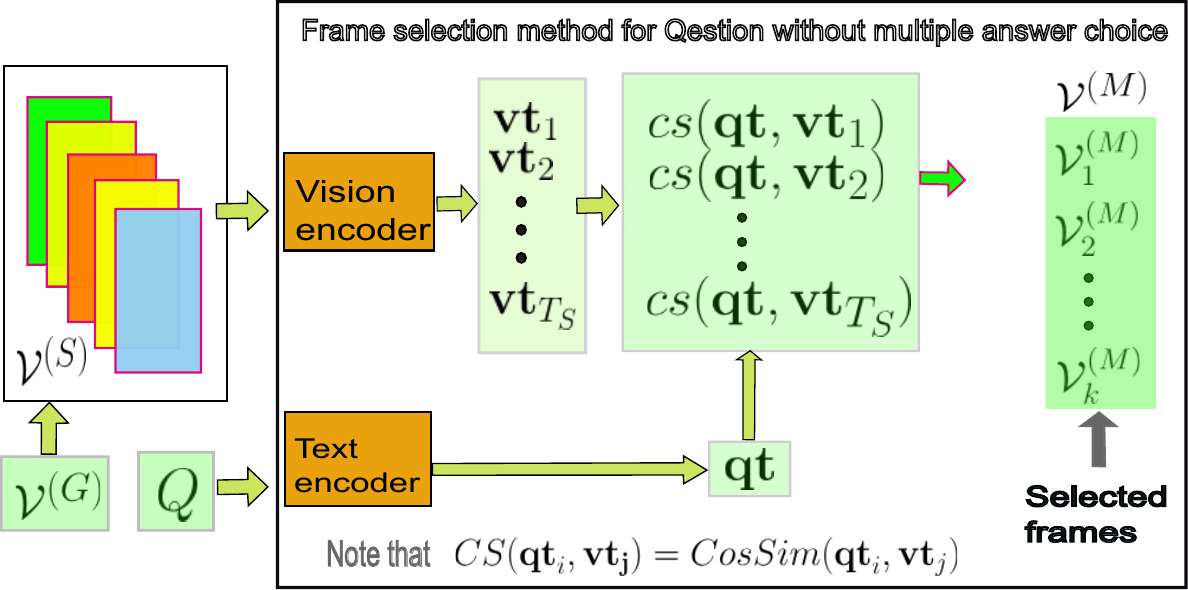}
    \caption{
    Proposed query-video (QV) aligned frame selection method using only the video and the question. First, to reduce computational and storage requirements, a subset $\mathcal{V}^{(S)}$ containing $T_s$ frames is sampled from the original video $\mathcal{V}^{(G)}$. These $T_s$ frames and the question $Q$ are then encoded into vector representations using a vision-language encoder (e.g., CLIP or LongCLIP). Next, the $k$ frames with the highest cosine similarity scores between the video-frame and text-query vectors are selected to form the optimal subset $\mathcal{V}^{(M)}$. The value of $k$ is typically determined by the target MLLM and its input-length constraint.
    }
    \label{fig:QVframework}
\end{figure}
\par
Note that these vision-language models process image and language inputs separately to produce token representations that encode semantic information from each modality. An image or text token representation can be viewed as an abstract vector representation of the input content. If an image and a text query share related semantic information, their vector representations tend to be aligned in the embedding space. For example, consider a video frame showing a person working with drawings at a construction site and a text question asking about the profession of the person working with drawings in the video. In this case, the image token vector $\textbf{vt}$ of the video frame and the text token vector $\textbf{qt}$ of the question will exhibit strong alignment; consequently, their $CosSim(\textbf{vt}, \textbf{qt})$ similarity score will be high. By evaluating the alignment between each video frame and the text question, the frame selection method can identify frames that are more relevant to the question, as illustrated in Fig.~\ref{fig:EffectOfVideoFrameSelection}.

\par
\begin{definition} 
Let a vision-language model $\mathcal{VL}_{\theta_{vl}}(\cdot)$ produce a token vector $\mathbf{vt} = \mathcal{VL}_{\theta_{vl}}(\mathbf{V})$ from a video frame $\mathbf{V}$ and a text token vector $\mathbf{qt} = \mathcal{VL}_{\theta_{vl}}(Q)$ from a text query $Q$. We define
 \begin{equation}
   R_{\mathcal{M}_{{\theta}_{vl}}}(Q,\mathbf{V} ) = CosSim( \mathbf{qt},\mathbf{vt})
 \end{equation}   
to estimate the usefulness of $\mathbf{V}$ for answering the question $Q$ with an MLLM.
   \label{eq:untilityVL}
\end{definition}

Using Def.~\ref{eq:untilityVL} to estimate the usefulness of a video frame $\mathbf{V}$ for answering a text question $Q$, we define a frame subset selector $FS^{(q)}(\cdot,\cdot)$ that aims to improve the performance of a downstream MLLM. Let the
\begin{equation}
FS^{(q)}(Q,\mathcal{V}^{(S)}) = \underset{[k] \subseteq [T_s]} {\arg \max} \sum_{i\in [k]} R_{\mathcal{M}_{\theta_{m}}}(Q,\mathbf{V}^{(S)}_i)
\label{eq:QueryVideo}
\end{equation}

Figure~\ref{fig:QVframework} provides an overview of the computational steps required to implement the function defined in Eq.~\ref{eq:QueryVideo}. A program-like description of these steps is presented in Algorithm~\ref{alg:qv-selection} in the next section. The algorithm evaluates every frame $\mathbf{V}^{(S)}_i \in \mathcal{V}^{(S)}$ and then selects the most promising $k$ frames.
From the algorithm, it is easy to see that the computational complexity of the function $FS^{(q)}(Q,\mathcal{V}^{(S)})$, which selects $k$ frames from the $T_s$ video frames in $\mathcal{V}^{(S)}$, is $O(T_s)$. This is because evaluating each frame takes constant time, while selecting the $k$ best frames from the $T_s$ candidate frames requires $O(k\log_2 T_s)$ time. We assume that $k\log_2(T_s) \leq T_s$.

\subsection{Query-Answer-Video Aligned Frame Selection}
\label{sec:QueryAnswerAligned}
The QV method described above selects frames only based on the question, but the frame-selection process can be improved when the question provides a set of possible answers, i.e., answer choices. In that case, we can pair each answer $a_i$ with the question to have $n_a$ question-answer pairs, where $n_a$ is the number of answer choices. We define a function for selecting the most promising $k$ frames:
\begin{equation}
FS^{(qa)}(QA,\mathcal{V}^{(S)}) = \underset{[k] \subseteq [T_s]} {\arg \max} \sum_{i\in [k]} \left( \max_{j\in [n_a]} R_{\mathcal{M}_{\theta_{m}}}(Qa_j,\mathbf{V}^{(S)}_i) \right)
\label{eq:QueryAnswerVideo}
\end{equation}

\begin{figure}[htb]
    \centering
    \includegraphics[width=.95\textwidth]{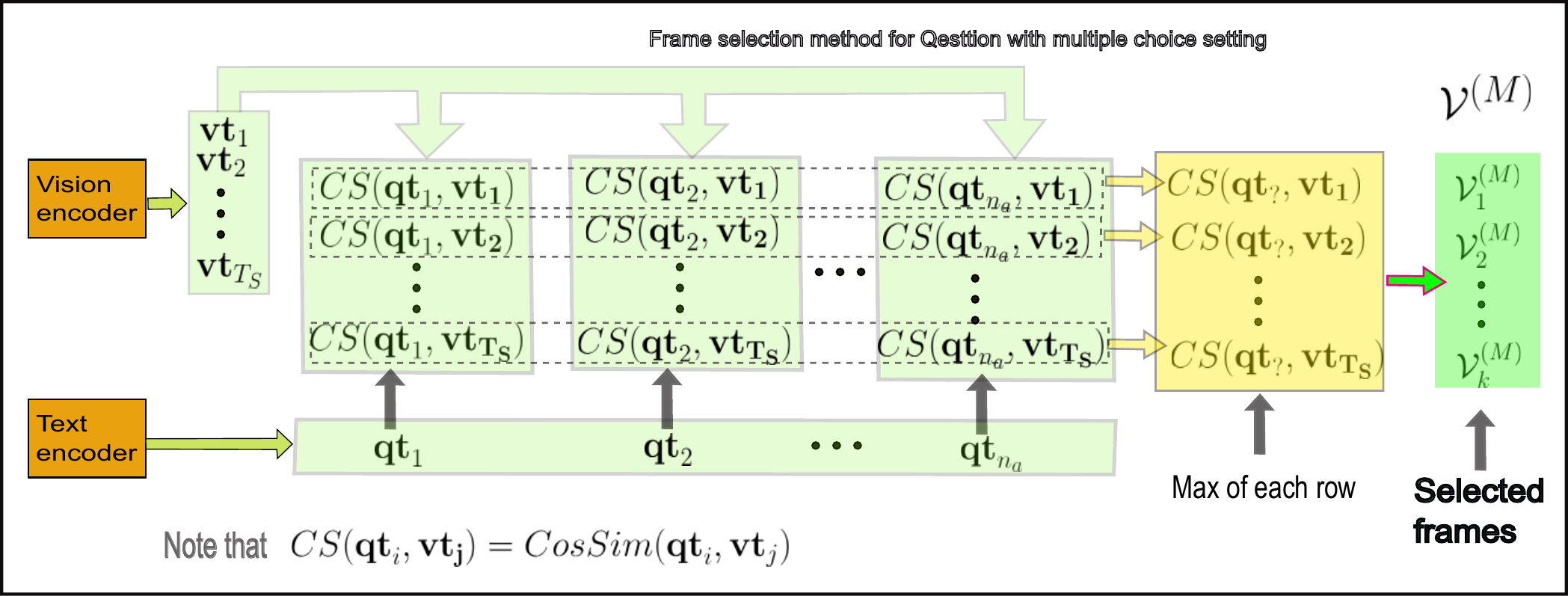}
    \caption{Proposed query-answer-video (QAV)-aligned frame-selection method for questions accompanied by a set of answer choices, commonly known as multiple-choice questions. This figure illustrates only the computational steps performed after the frame and text vectors have been generated by the encoders. Because the question is associated with $n_a$ answer choices, we generate $n_a$ question-answer-pair vectors. The cosine similarity between each question-answer vector and each frame vector is then computed, producing one column of similarity scores for each $qt_i$. After computing all $n_a$ columns, the highest score in each row is selected and stored in a separate column (highlighted in yellow). Finally, the $k$ frames with the highest alignment scores are selected.}
    \label{fig:QAVframework}
\end{figure}

A computationally efficient implementation of the function above is similar to that of $FS^{(q)}(Q,\mathcal{V}^{(S)})$ in Eq.~\ref{eq:QueryVideo}, but requires a preliminary step. As shown in Fig.~\ref{fig:QAVframework}, each frame $\mathbf{V}^{(S)}_i \in \mathcal{V}^{(S)}$ is evaluated $n_a$ times, one for each $Qa_j$, and then the maximum value from these $n_a$ values is assigned to the frame, as shown in the figure's yellow column. In the next section, Algorithm~\ref{alg:qav-selection} presents a program-like outline of the steps involved in the proposed method. As shown in the rightmost column of the figure, after computing all scores for all $T_s$ frames, the most promising $k$ frames are selected. Thus, the computational complexity of the function $FS^{(qa)}(Q,\mathcal{V}^{(S)})$ for selecting $k$ video frames from a set of $T_s$ video frames is $O(n_aT_s)$, because we evaluate each of the $T_s$ frames in $O(n_a)$ time, and selecting the best $k$ candidate frames takes $O(k\log_2 T_s)$ time. As before, it is assumed that $k\log_2 T_s \leq T_s$.

\subsection{}

\subsection{Algorithms for Proposed Frame Selection}
\label{sec:appendix-algorithms}
The descriptions of the two proposed frame-selection methods in Sections~\ref{sec:QueryAligned} and~\ref{sec:QueryAnswerAligned} are presented in this section. Algorithm~\ref{alg:qv-selection} provides a program-like description of the query-video-frame-aligned frame-selection procedure, whereas Algorithm~\ref{alg:qav-selection} provides a program-like description of the query-answer-video-frame-aligned frame-selection procedure.
\begin{algorithm}[H]
\caption{Query-aligned video frame selection (QV)}
\label{alg:qv-selection}
\begin{algorithmic}[1]
\Require Video $\mathcal{V}^{(G)}$, question $Q$, frame budget $k$,
         sampling interval $\Delta t{=}1\,\mathrm{s}$,
         vision-language encoder $\mathcal{VL}_{\theta_{vl}}$
\Ensure Chronologically ordered selected frame sequence $\mathcal{V}^{(M)}$

\State $\mathcal{V}^{(S)} \gets \Call{SampleFrames}{\mathcal{V}^{(G)},\,\Delta t}$
       \Comment{1 frame per second}
\State $T_s \gets |\mathcal{V}^{(S)}|$

\State $\mathbf{qt} \gets
    \mathcal{VL}^{\mathrm{text}}_{\theta_{vl}}(Q)$

\For{$i = 1$ \textbf{to} $T_s$}
    \State $\mathbf{vt}_i \gets
        \mathcal{VL}^{\mathrm{image}}_{\theta_{vl}}
        (\mathbf{V}^{(S)}_i)$
    \State $s_i \gets \operatorname{CosSim}(\mathbf{qt},\,\mathbf{vt}_i)$
\EndFor

\State $\mathcal{H} \gets \Call{BuildMaxHeap}{\{(s_i,i)\}_{i=1}^{T_s}}$
\State $\mathcal{I} \gets \emptyset$
\For{$r = 1$ \textbf{to} $k$}
    \State $(s,i) \gets \Call{ExtractMax}{\mathcal{H}}$
    \State $\mathcal{I} \gets \mathcal{I} \cup \{i\}$
\EndFor

\State $\mathcal{I} \gets \Call{SortAscending}{\mathcal{I}}$
       \Comment{restore temporal order}
\State \Return $\mathcal{V}^{(M)} =
       \bigl(\mathbf{V}^{(S)}_i \mid i \in \mathcal{I}\bigr)$
\end{algorithmic}
\end{algorithm}

These procedures were used in our experiments. Both procedures first construct a compact candidate set by subsampling the input video at a fixed rate, set to $1$\,fps in our experiments, score each candidate frame with a pretrained vision-language encoder, select the top-$k$ candidates by score, and restore the selected frames to their original temporal order before passing them to the downstream MLLM. $CosSim$ is defined in Eq.~\ref{eq:cosSim} and already includes norm-based normalization.

\begin{algorithm}[H]
\caption{Query-answer-aligned video frame selection (QAV)}
\label{alg:qav-selection}
\begin{algorithmic}[1]
\Require Video $\mathcal{V}^{(G)}$, question $Q$,
         answer choices $A = \{a_j\}_{j=1}^{n_a}$,
         frame budget $k$,
         sampling interval $\Delta t{=}1\,\mathrm{s}$,
         vision-language encoder $\mathcal{VL}_{\theta_{vl}}$
\Ensure Chronologically ordered selected frame sequence $\mathcal{V}^{(M)}$

\State $\mathcal{V}^{(S)} \gets \Call{SampleFrames}{\mathcal{V}^{(G)},\,\Delta t}$
       \Comment{1 frame per second}
\State $T_s \gets |\mathcal{V}^{(S)}|$

\For{$j = 1$ \textbf{to} $n_a$}
    \State $Q\!a_j \gets Q \,\|\, a_j$
           \Comment{concatenate question with $j$-th answer choice}
    \State $\mathbf{qt}_j \gets
        \mathcal{VL}^{\mathrm{text}}_{\theta_{vl}}
        (Q\!a_j)$
\EndFor

\For{$i = 1$ \textbf{to} $T_s$}
    \State $\mathbf{vt}_i \gets
        \mathcal{VL}^{\mathrm{image}}_{\theta_{vl}}
        (\mathbf{V}^{(S)}_i)$
    \For{$j = 1$ \textbf{to} $n_a$}
        \State $S_{ij} \gets \operatorname{CosSim}(\mathbf{qt}_j,\,\mathbf{vt}_i)$
    \EndFor
    \State $s_i \gets \max_{j \in [n_a]} S_{ij}$
           \Comment{best alignment across all choices}
\EndFor

\State $\mathcal{H} \gets \Call{BuildMaxHeap}{\{(s_i,i)\}_{i=1}^{T_s}}$
\State $\mathcal{I} \gets \emptyset$
\For{$r = 1$ \textbf{to} $k$}
    \State $(s,i) \gets \Call{ExtractMax}{\mathcal{H}}$
    \State $\mathcal{I} \gets \mathcal{I} \cup \{i\}$
\EndFor

\State $\mathcal{I} \gets \Call{SortAscending}{\mathcal{I}}$
       \Comment{restore temporal order}
\State \Return $\mathcal{V}^{(M)} =
       \bigl(\mathbf{V}^{(S)}_i \mid i \in \mathcal{I}\bigr)$
\end{algorithmic}
\end{algorithm}


\section{Experimental Evaluation}
\label{sec:evaluation}
We evaluate two versions of our method (QV and QAV) in comparison to uniform sampling and three training-free frame selection baselines on three benchmarks and three open-source MLLMs.
\subsection{Details of Experimental Setup}
Our experiments use three benchmark datasets that all other known frame-selection methods have used; we discovered that the questions in these datasets happen to be in the multiple-choice question-answering format. Given a video $(V)$, a question $(Q)$, and answer choices $(A)$, a frame-selection method is used to select $k$ frames from the video. These frames are kept in their original temporal order and then sent to the downstream MLLM along with the question and answer choices. Unless stated otherwise, we use $k = 16$ frames for uniform sampling, AKS, FOCUS, and our methods. For Q-Frame, we follow the frame budget specified in the original paper, which selects $4 + 8 + 32$ frames across multiple resolutions. We use accuracy as the primary evaluation metric.

\subsubsection{Benchmark Datasets and MLLMs}
\label{sec:Dataset}
We evaluate three public benchmarks (\textbf{MLVU}~\cite{MLLVUbenchmarkZhou2024_mlvu}, 
\textbf{Video-MME}~\cite{fu2024video_mme}, and \textbf{LongVideoBench}~\cite{LongVideoBenchWu2024})   for long-video multiple-choice question answering. These datasets cover different video durations, domains, and reasoning skills.

\paragraph{MLVU benchmark}
\label{sec:appendix-mlvu-subcategories}

\begin{figure}[H]
    \centering
    \includegraphics[width=0.75\linewidth]{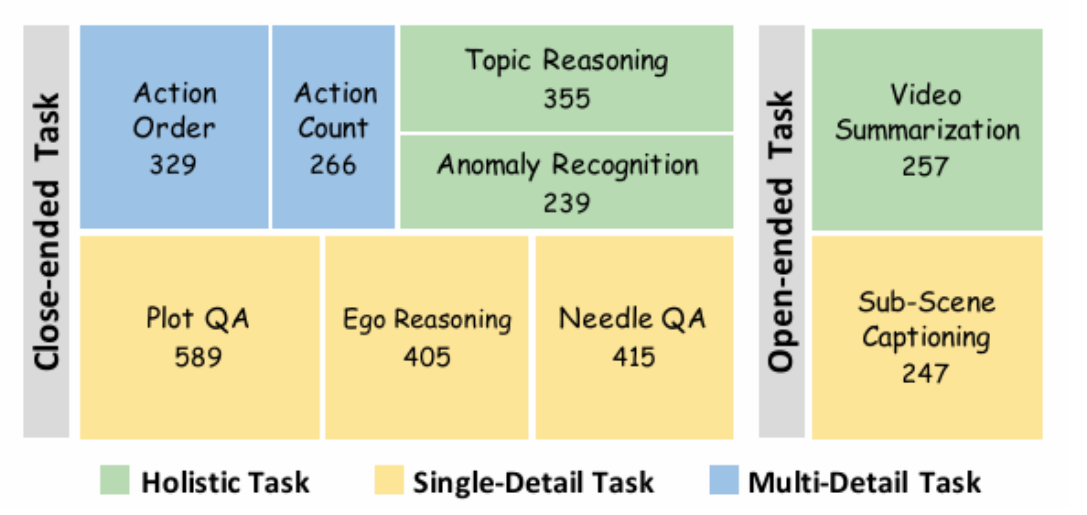}
    \caption{This is part of Figure 1 in the MLVU benchmark paper~\cite{zhou2024mlvu}. It shows nine task subcategories. Our study evaluates only the seven closed-ended tasks.}
    \label{fig:MLVUbenchSubCategories}
\end{figure}

As shown in Fig.~\ref{fig:MLVUbenchSubCategories}, \textbf{MLVU}~\cite{MLLVUbenchmarkZhou2024_mlvu} consists of long videos that cover several video understanding tasks. We evaluate the seven closed-ended categories: Topic Reasoning (TR), Anomaly Recognition (AR), Needle QA (NQA), Ego Reasoning (ER), Plot QA (PQA), Action Order (AO), and Action Count (AC).

\begin{table}[H]
\centering
\caption{MLVU task subcategories and number of questions.}
\label{tab:mlvu-subcategories}
\begin{tabular}{cccccccccc}
\toprule
& \multicolumn{7}{c}{Closed-Ended Task} & \multicolumn{2}{c}{Open-Ended Task} \\
\cmidrule(lr){2-8} \cmidrule(lr){9-10}
Task & TR & AR & NQA & ER & PQA & AO & AC & VS & SSC \\
\midrule
Questions & 355 & 239 & 415 & 405 & 589 & 329 & 266 & 257 & 247 \\
\bottomrule
\end{tabular}
\end{table}

\paragraph{Video-MME benchmark} \textbf{Video-MME}~\cite{fu2024video_mme} consists of short-, medium-, and long-duration videos spanning a wide range of domains. The benchmark is designed to evaluate the video-understanding and reasoning capabilities of multimodal large language models across videos of varying lengths and complexity. During question answering, models can be evaluated either with access to the video subtitles or without subtitle information. Since subtitles often provide complementary textual cues that can facilitate better understanding, we evaluate and report results under both settings, namely with subtitles and without subtitles, to provide a comprehensive evaluation of model performance.

\paragraph{LongVideoBench benchmark} \textbf{LongVideoBench}~\cite{LongVideoBenchWu2024} focuses on the understanding of long-context videos with diverse durations and content types. The benchmark contains questions that often require reasoning across multiple video segments and the integration of information distributed throughout the video. To evaluate the impact of  textual description of scenes, models can be tested with or without access to video subtitles. We we have evaluated and report results for LongVideoBench under both subtitle-free and subtitle-augmented settings to provide a comprehensive assessment of model performance.

\paragraph{MLLM Models}
\label{sec:MLLM Models}
We evaluate three open-source MLLMs: \textbf{LLaVA-Mini-8B}~\cite{LLaVAMiniZhang2025llava}, \textbf{Qwen2-VL-7B}~\cite{Qwen2VLwang2024qwen2_vl}, and \textbf{LLaVA-Video-7B}~\cite{zhang2024llava-video}. LLaVA-Mini is included to evaluate whether query-guided frame selection can benefit a compact MLLM. In contrast, Qwen2-VL and LLaVA-Video provide a comparison against prior frame-selection methods, as they represent stronger 7B-scale video MLLMs. All models are evaluated without fine-tuning or architectural modifications. Therefore, any performance differences arise solely from the frames selected and provided to the downstream model.
\subsubsection{Implementation Details,  Compute Resources, and Fair Comparison Protocol }
\paragraph{Implementation Details}
\label{sec:implemenation}
For each video, we begin by constructing a candidate frame pool $\mathcal{V}^{(S)}$ by sampling frames at a fixed temporal interval. In the QV variant, each candidate frame is encoded into a visual embedding, while the question $Q$ is encoded into a text embedding. We then score each frame based on the cosine similarity between its visual embedding and the question embedding. The top-$k$ frames with the highest scores are selected and then restored to their original temporal order before being passed to the downstream MLLM. For the QAV variant, we additionally incorporate the answer choices as semantic cues. Each answer choice is paired with the question to form a set of question-answer text inputs. Each pair is encoded separately. Each candidate frame is scored by its maximum cosine similarity over all question-answer pairs. The top-$k$ frames under this score are selected and reordered chronologically. Importantly, no gradient updates, model fine-tuning, or dataset-specific training are used for any of the proposed frame selectors.

\paragraph{Compute Resources}
\label{sec:compute-resources}
Experiments were conducted on a Linux server with four NVIDIA H100 NVL GPUs, each with 96 GB of GPU memory, dual Intel Xeon Silver 4514Y CPUs, and 1.5 TB of system memory. Each evaluation job used one H100 GPU, and independent jobs were run in parallel when multiple GPUs were available. No model training or finetuning was performed; compute was used only for frame selection and MLLM inference. A full benchmark evaluation took approximately 4 to 24 GPU-hours, depending on the MLLM, benchmark, frame budget, and subtitle setting.

\paragraph{Fair Comparison Protocol} 
For a fair comparison, we evaluate the frame-selection methods under the same downstream MLLM and benchmark settings whenever possible. Specifically, we 
\begin{inparaenum}[$(i)$] 
\item maintain a fixed budget of $k=16$ selected frames for uniform sampling and for AKS, FOCUS, and our QV/QAV methods, 
\item test Q-Frame using its released multi-resolution setting ($4+8+32$ frames) and, when available, an additional 16-frame reproduction, 
\item use the  BLIP-based frame selections for AKS and FOCUS, and use CLIP-family encoders (CLIP or LongCLIP) for Q-Frame and our methods, 
\item report the same seven closed-ended MLVU task categories for all MLVU comparisons, and 
\item provide all duration splits, both with and without subtitles, for Video-MME and LongVideoBench.
\end{inparaenum} 
We base our evaluations on the released evaluation code of Q-Frame, AKS, and FOCUS whenever possible, making only the necessary adaptations to run the same benchmarks, model backbones, and local data paths.

\subsection{Main Results}
\label{sec:MainResults}

\subsubsection{Comparison with SOTA Methods}
\label{sec:SotaComparisons}
As stated before, we compare against three recent training-free frame-selection baselines: Q-Frame (CVF 2025)~\cite{QframeZhang2025q}, AKS (CVPR 2025)~\cite{tang2025adaptive}, and FOCUS (ICLR 2026)~\cite{zhu2026focus}. Because these are the most recent methods and have demonstrated significantly improved performance, we compare and contrast our proposed methods with them. These methods have reported \textit{only} average performance across question categories for the MLVU benchmark. We provide more fine-grained evaluations by reporting question-category-wise performance in addition to average performance. For Video-MME and LongVideoBench, we report results both with and without subtitles, which were not reported in earlier works.
\par
\paragraph{MLVU benchmark} 
We started our evaluation with three MLLMs, including LLaVA-Mini. As shown in Table~\ref{tab:mlvu_results}, LLaVA-Mini achieves lower overall video-understanding accuracy than Qwen2-VL and LLaVA-Video. Its most distinctive pattern is its Topic Reasoning (TR) score. The base model reaches 76.0\% on TR, which is 26.0 percentage points higher than its next-highest subcategory score, Anomaly Recognition (AR), at 50.0\%. For Qwen2-VL, QAV with LongCLIP achieves the best average accuracy of 69.2\%. Among the prior frame-selection baselines, FOCUS is the closest competitor at 66.19\%, so QAV improves upon it by 3.0 percentage points. For LLaVA-Video, QAV with LongCLIP also achieves the best average accuracy of 71.3\%. This is 0.9 percentage points higher than Q-Frame, which achieves 70.4\% using its released multi-resolution setting of 44 frames (4 + 8 + 32). FOCUS is the strongest prior 16-frame baseline at 69.0\%, which is 2.3 percentage points lower than QAV.

 \begin{table}[htb]
\caption{Comparison of different methods on MLVU. QV = Query-Video frame aligned and QAV = Query-Answer-Video Frame aligned.
Best performance is boldfaced, and the second-best is underlined.}
\small
\setlength{\tabcolsep}{5pt}
\renewcommand{\arraystretch}{1.08}
\resizebox{\textwidth}{!}{%
\begin{tabular}{llccccccccc}
\toprule
Methods & Size & \#Frames & TR & AR & NQA & ER & PQA & AO & AC & Avg. \\
\specialrule{0.08em}{0em}{0em}
Video-ChatGPT & 7B & 100 & 26.9 & 24.0 & 40.3 & 42.0 & 29.9 & 25.1 & 31.1 & 31.3 \\
MovieChat & 7B & 2048 & 29.5 & 25.0 & 24.2 & 24.7 & 25.8 & 28.6 & 22.8 & 25.8 \\
Movie-LLM & 7B & 1fps & 30.0 & 29.0 & 29.6 & 24.7 & 24.1 & 20.5 & 24.8 & 26.1 \\
TimeChat & 7B & 96 & 23.1 & 27.0 & 24.5 & 28.4 & 25.8 & 24.7 & \textbf{32.0} & 30.9 \\
LLaMA-VID & 7B & 1fps & 50.8 & 34.5 & 30.1 & 32.7 & 32.5 & 23.9 & 27.8 & 33.2 \\
MA-LMM & 7B & 1000 & 51.9 & 35.5 & 43.1 & 38.9 & 35.8 & 25.1 & 24.3 & 36.4 \\

\midrule
LLaVA-Mini & 8B & 1fps & \textbf{76.0} & 50.0 & 44.5 & 37.5 & 49.0 & 24.3 & 18.4 & 42.8 \\
LLaVA-Mini + QV(Ours) CLIP & 8B & 16 & 68.1 & \underline{51.5} & \textbf{71.6} & \textbf{47.7} & \underline{59.4} & \textbf{25.9} & \underline{30.6} & \underline{53.1} \\
LLaVA-Mini + QAV (Ours) CLIP & 8B & 16 & \underline{74.9} & \textbf{56.5} & \underline{70.7} & \underline{46.0} & \textbf{63.3} & 24.3 & \textbf{31.6} & \textbf{54.8} \\

\midrule
Qwen2-VL (Uniform) & 7B & 16 & \textbf{86.7} & \textbf{71.5} & 72.4 & 56.3 & 60.9 & 42.9 & 22.8 & 60.4 \\
Qwen2-VL + Q-Frame & 7B & 4+8+32 & 76.8 & 57.0 & 74.6 & 48.0 & 59.9 & 40.5 & 22.3 & 56.3 \\
Qwen2-VL + FOCUS  & 7B & 16 & 80.99 & 65.00 & 81.13 & 62.78 & 69.76 & 47.88 & \textbf{41.75} & 66.19 \\
Qwen2-VL + AKS & 7B & 16 & 82.1 & 64.0 & 78.9 & 56.8 & 63.3 & 44.4 & \underline{40.3} & 62.7 \\
Qwen2-VL + QV + LongCLIP (Ours) & 7B & 16 & \underline{84.4} & 67.5 & \textbf{87.0} & 65.1 & 71.6 & 50.6 & 28.6 & 67.7 \\
Qwen2-VL + QAV+ CLIP (Ours) & 7B & 16 & 82.1 & \underline{70.0} & 86.1 & \underline{66.7} & \underline{74.4} & \underline{54.4} & 28.6 & \underline{68.8} \\
Qwen2-VL + QAV + LongCLIP (Ours) & 7B & 16 & 81.0 & \underline{70.0} & \underline{86.2} & \textbf{67.6} & \textbf{75.0} & \textbf{55.2} & 29.6 & \textbf{69.2} \\
\midrule
LLaVA-Video + Q-Frame & 7B & 4+8+32 & \textbf{82.9} & 52.0 & 81.4 & 68.5 & \underline{77.4} & \textbf{59.8} & \textbf{51.9} & \underline{70.4} \\
LLaVA-Video + FOCUS & 7B & 16 & 81.0 & \textbf{59.0} & \underline{83.4} & 69.0 & 73.8 & 49.8 & \underline{49.5} & 69.0 \\
LLaVA-Video + AKS  & 7B & 16 & 82.1 & \underline{57.5} & 77.7 & 61.9 & 70.3 & 45.9 & 48.5 & 65.5 \\
LLaVA-Video + QV+ LongCLIP (Ours) & 7B & 16 & 81.7 & 50.5 & \textbf{84.5} & \underline{70.7} & 75.0 & 49.4 & 47.6 & 68.8 \\
LLaVA-Video + QAV+ LongCLIP (Ours) & 7B & 16 & 79.1 & 57.0 & \textbf{84.5} & \textbf{71.0} & \textbf{80.1} & \underline{57.1} & 48.1 & \textbf{71.3} \\
\bottomrule
\end{tabular}%
}
\label{tab:mlvu_results}
\end{table}
\par
\paragraph{Video-MME benchmark} 
Video-MME supports evaluation with and without subtitles. As shown in Table~\ref{tab:videomme}, we evaluate all methods under both settings. On Qwen2-VL, QAV with LongCLIP achieves the best overall accuracy in both cases. Without subtitles, it achieves 60.90\%, outperforming the strongest prior baseline, FOCUS, by 3.83 percentage points. With subtitles, it achieves 62.93\%, outperforming the strongest prior baseline, Q-Frame, by 2.04 percentage points. On LLaVA-Video, AKS is marginally better than QAV without subtitles by 0.12 percentage points (61.93\% vs. 61.81\%). With subtitles, however, QAV achieves the best overall accuracy, outperforming the strongest prior baseline, Q-Frame, by 0.81 percentage points (64.70\% vs. 63.89\%).

\begin{table}[H]
\centering
\caption{Video-MME results (without / with subtitles). Best performance is boldfaced and 2nd best is underlined.}
\label{tab:videomme}
\setlength{\tabcolsep}{3pt}
\resizebox{\textwidth}{!}{%
\begin{tabular}{l c c cccc}
\toprule
\multirow{2}{*}{\textbf{Model}} & \textbf{LLM} & \multirow{2}{*}{\textbf{\#Frames}} & \multicolumn{4}{c}{\textbf{Video-MME} (\underline{wo\,/\,w subs})} \\
\cmidrule(lr){4-7}
 & \textbf{Size} & & Overall & Short & Medium & Long \\
 & & & \underline{17min} & \underline{1.3min} & \underline{9min} & \underline{41min} \\
\midrule
\multicolumn{7}{l}{\underline{Fixed input tokens:}} \\
Qwen2-VL + Uniform & 7B & 16 & 56.07\,/\,57.44 & 68.22\,/\,68.30 & 52.33\,/\,55.60 & 47.67\,/\,48.40 \\
Qwen2-VL + Q-Frame  & 7B & 4+8+32 & 56.96\,/\,60.89 & 68.33\,/\,\underline{72.33} & \underline{55.44}\,/\,59.11 & 47.11\,/\,51.22 \\
Qwen2-VL + AKS & 7B & 16 & 55.19\,/\,54.70 & 64.44\,/\,64.11 & 53.22\,/\,53.00 & 47.89\,/\,47.00 \\
Qwen2-VL + FOCUS & 7B & 16 & 57.07/59.33 & 67.56/68.44 & 54.44/56.67 & \underline{49.22}/52.89 \\

Qwen2-VL + QV (Ours)  & 7B & 16 & \underline{57.11}\,/\,\underline{61.04} & \underline{68.78}\,/\,70.22 & 54.33\,/\,\underline{59.89} & 48.22\,/\,\underline{53.00} \\
Qwen2-VL + (QAV + LongCLIP (Ours)) & \textbf{7B} & \textbf{16} & \textbf{60.9}\,/\,\textbf{62.93} & \textbf{70.7}\,/\,\textbf{72.67} & \textbf{60.8}\,/\,\textbf{62.56} & \textbf{51.1}\,/\,\textbf{53.56} \\
\bottomrule
LLaVA-Video (Uniform) & 7B & 16 & 59.56/62.07 & 71.33/74.11 & 58.22/60.22 & 49.11/51.89 \\
\textcolor{orange}{
LLaVA-Video (+Q-Frame)} & \textcolor{orange}{7B} & \textcolor{orange}{16} & \textcolor{orange}{56.30\,/\,58.56 }& \textcolor{orange}{65.67\,/\,66.60} & \textcolor{orange}{53.56\,/\,56.80 }& \textcolor{orange}{49.67\,/\,52.30} \\
LLaVA-Video (+Q-Frame) & 7B & 4+8+32 & 60.44\,/\,\underline{63.89} & 72.22\,/\,74.33 & 57.89\,/\,\underline{62.00} & 51.22\,/\,\textbf{55.33} \\
LLaVA-Video (+AKS) & 7B & 16 & \textbf{61.93}\,/\,63.11 & \underline{73.11}\,/\,\underline{74.67} & \underline{60.89}\,/\,61.00 & \underline{51.78}\,/\,53.67 \\
LLaVA-Video (+FOCUS) & 7B & 16 & 60.52\,/\,60.70 & 70.40\,/\,69.70 & 58.80\,/\,59.00 & \textbf{52.30}\,/\,53.40 \\
LLaVA-Video + QV (Ours) & 7B & 16 & 58.07\,/\,62.15 & 69.56\,/\,72.78 & 56.33\,/\,60.78 & 48.33\,/\,52.89 \\
LLaVA-Video + (QAV + LongCLIP (Ours)) & 7B & 16 & \underline{61.81}\,/\,\textbf{64.70} & \textbf{73.22}\,/\,\textbf{76.67} & \textbf{61.11}\,/\,\textbf{63.56} & 51.11\,/\,\underline{53.89} \\
\bottomrule
\end{tabular}%
}
\end{table}

\paragraph{LongVideoBench} 

LongVideoBench supports evaluation with and without subtitles. As shown in Table~\ref{tab:longvideobench}, we report both settings for all methods. On Qwen2-VL, QAV with LongCLIP achieves the best overall accuracy in both cases. Without subtitles, it obtains 58.80\%, outperforming the strongest prior baseline, FOCUS, by 0.76 percentage points. With subtitles, it obtains 60.20\%, outperforming the strongest prior baseline, FOCUS, by 2.01 percentage points. On LLaVA-Video, QAV with LongCLIP also achieves the best overall accuracy in both settings. Without subtitles, it obtains 61.32\%, improving over the strongest prior baseline, AKS, by 1.11 percentage points. With subtitles, it obtains 63.49\%, improving over the strongest prior baseline, Q-Frame, by 2.98 percentage points.

\par
In summary, the proposed QV and QAV methods are usually the strongest or among the strongest methods across the three benchmarks. Among the prior frame-selection baselines, FOCUS is most often the closest competitor in the category-wise and video-duration-wise comparisons reported in Tables~\ref{tab:mlvu_results}, \ref{tab:videomme}, and \ref{tab:longvideobench}. These results suggest that query- and answer-aware frame selection provides a robust alternative to uniform sampling and recent training-free frame-selection baselines under a fixed frame budget.

\begin{table}[H]
\centering
\caption{LongVideoBench results (without / with subtitles). Best performance is boldfaced and 2nd best is underlined.}
\label{tab:longvideobench}
\setlength{\tabcolsep}{3pt}
\resizebox{\textwidth}{!}{%
\begin{tabular}{l c c ccccc}
\toprule
\multirow{2}{*}{\textbf{Model}} & \textbf{LLM} & \multirow{2}{*}{\textbf{\#Frames}} & \multicolumn{5}{c}{\textbf{LongVideoBench} (\textit{wo\,/\,w subs})} \\
\cmidrule(lr){4-8}
 & \textbf{Size} & & Overall & 8s--15s & 15s--60s & 3m--10m & 15m--60m \\
\midrule
Qwen2-VL (+Uniform) & 7B & 16 & 54.10\,/\,55.80 & \underline{66.70}\,/\,65.10 & 65.10\,/\,68.00 & 52.70\,/\,57.50 & 47.50\,/\,47.70 \\
Qwen2-VL (+ FOCUS) & 7B & 16 & 58.04\,/\,58.19 & 66.14\,/\,\textbf{67.73} & 66.28\,/\,66.86 & 56.07\,/\,58.01 & \textbf{54.26}\,/\,52.48 \\
Qwen2-VL (+AKS) & 7B & 16 & 54.90\,/\,54.45 & 62.43\,/\,60.85 & 61.63\,/\,62.21 & 56.55\,/\,54.85 & 49.11\,/\,49.65 \\
Qwen2-VL + (Q-Frame)  & 7B & 4+8+32 & 57.29/57.44 & 62.96\,\,/63.49 & \textbf{73.26}\,/\,69.19 & \underline{58.25}/\underline{59.47} & 49.82/50.35 \\
Qwen2-VL (+ QV + LongCLIP (Ours)) & \textbf{7B} & \textbf{16} & \underline{58.12}\,/\,\underline{59.01} & \textbf{68.25}\,/\,\underline{65.61}& 69.19\,/\,\underline{70.93} & 57.52\,/\,\underline{59.47} & \underline{51.77}\,/\,\underline{52.84} \\
Qwen2-VL (+ QAV + LongCLIP (Ours)) & \textbf{7B} & \textbf{16} & \textbf{58.80}\,/\,\textbf{60.20} & 65.60\,/\,65.10 & \underline{69.80}\,/\,\textbf{72.7} & \textbf{59.70}\,/\,\textbf{61.20} & \underline{52.50}\,/\,\textbf{54.10} \\

\bottomrule
LLaVA-Video (+Uniform) & 7B & 16 & 56.90\,/\,59.30 & \underline{67.20}\,/\,69.31 & 70.35\,/\,72.09 & 53.64\,/\,57.28 & 51.69\,/\,53.48 \\
\textcolor{orange}{LLaVA-Video (+Q-Frame)} & \textcolor{orange}{7B} & \textcolor{orange}{16} & \textcolor{orange}{55.80\,/\,56.25 }& \textcolor{orange}{\textbf{70.90}\,/\,67.73} & \textcolor{orange}{66.28\,/\,68.61} & \textcolor{orange}{52.67\,/\,53.88} & \textcolor{orange}{49.82\,/\,50.36 }\\
LLaVA-Video (+Q-Frame) & 7B & 4+8+32 & 59.98\,/\,60.51 & \underline{67.20}\,/\,66.67 & 70.93\,/\,70.35 & 57.28\,/\,58.74 & \underline{56.21}\,/\,56.74 \\
LLaVA-Video (+FOCUS) & 7B & 16 & 58.86\,/\,60.06 & 62.96\,/\,64.55 & 68.61\,/\,67.44 & 58.25\,/\,59.71 & 54.97\,/\,56.56 \\
LLaVA-Video (+AKS) & 7B & 16 & \underline{60.21}\,/\,59.69 & \textbf{70.90}\,/\,67.72 & 69.77\,/\,73.26 & \underline{58.74}\,/\,58.01 & 54.79\,/\,54.08 \\
LLaVA-Video (+QV + LongCLIP (Ours)) & 7B & 16 & 59.99\,/\,\underline{62.83} & \underline{67.20}\,/\,\textbf{69.31} & \textbf{73.26}\,/\,\textbf{75.00} & 58.01\,/\,\underline{60.92} & 54.96\,/\,\textbf{58.33} \\
LLaVA-Video (+QAV + LongCLIP (Ours)) & 7B & 16 & \textbf{61.32}\,/\,\textbf{63.49} & \underline{67.20}/\,\textbf{69.31} & \underline{72.09}\,/\,\underline{73.84} & \textbf{59.71}\,/\,\textbf{63.59} & \textbf{57.22}\,/\,\underline{58.29} \\
\bottomrule
\end{tabular}%
}
\end{table} 
\subsubsection{Performance for Subcategories}
\label{sec:questionType}
\paragraph{MLVU benchmark} 
MLVU contains seven closed-ended and two open-ended task categories, as shown in Fig.~\ref{fig:MLVUbenchSubCategories} in Section~\ref{sec:Dataset}; we evaluate the seven closed-ended categories. On Qwen2-VL, uniform sampling performs best on the two holistic categories, TR and AR, exceeding the best proposed result by 2.3 and 1.5 percentage points, respectively. Our methods lead on NQA, ER, PQA, and AO, while FOCUS performs best on AC. On LLaVA-Video, Q-Frame leads on TR, AO, and AC; our methods lead on NQA, ER, and PQA; and FOCUS leads on AR. Note that Q-Frame uses its released multi-resolution setting of 44 frames (4 + 8 + 32), whereas FOCUS, AKS, uniform sampling, and our methods use 16 selected frames.

\subsubsection{Effect of Video Length}
\label{sec:video-length-effect}
\subparagraph{Video-MME benchmark} 
Video-MME has three duration groups and can be evaluated with or without subtitles, giving six settings. As shown in Table~\ref{tab:videomme}, our methods achieve the best results in all six settings on Qwen2-VL; among the prior baselines, Q-Frame is second-best in four settings and FOCUS in the remaining two. On LLaVA-Video, our methods lead in four short- and medium-video settings, while FOCUS and Q-Frame lead in one setting each.
\subparagraph{LongVideoBench}
LongVideoBench has four duration groups and can be evaluated with or without subtitles, giving eight settings. As shown in Table~\ref{tab:longvideobench}, our methods achieve the best results in five settings on Qwen2-VL, while FOCUS leads in two and Q-Frame in one. On LLaVA-Video, our methods lead or tie for the best result in seven settings; AKS and Q-Frame tie in the remaining setting.
\section{Ablation Study}
\label{sec:ablation}
 We have evaluated the effect of two VLM encoders, CLIP and LongCLIP, on the performance of our frame-selection methods. We have also evaluated the effect of the number of video frames on performance across all three benchmarks. For these studies, the LLaVA-Video MLLM with 7B parameters was used because, among the MLLMs we evaluated, it produced correct answers most often.
 
\subsection{Ablation Study for VLM Encodes}
\label{sec:appendix-encoder-ablation}
In this section, we report the effects of the number of video frames on all three benchmarks. We use LLaVA-Video-7B in our studies. Table~\ref{tab:encoder-ablation-mlvu} compares CLIP and LongCLIP as the vision-language encoders used in the proposed QV and QAV frame selectors. The downstream MLLM, dataset, and frame budget are kept constant. The results indicate that both selectors can operate effectively with either encoder. In this MLVU setting, LongCLIP provides modest average improvements for both QV and QAV, although the benefits vary across different task categories.

\begin{table}[H]
\centering
\caption{Encoder ablation for proposed frame selection on MLVU.
All rows use Qwen2-VL-7B as the downstream MLLM, with a fixed budget of $K=16$ selected frames. QV uses the question only; QAV uses max-over-options scoring with the question and answer choices.}
\label{tab:encoder-ablation-mlvu}
\small
\setlength{\tabcolsep}{5pt}
\renewcommand{\arraystretch}{1.08}
\resizebox{\textwidth}{!}{%
\begin{tabular}{l l c ccccccc c}
\toprule
Method & Encoder & \#Frames & TR & AR & NQA & ER & PQA & AO & AC & Avg. \\
\midrule
QV  & CLIP     & 16 & 83.3 & 66.0 & 87.2 & 65.1 & 71.1 & 49.4 & 28.6 & 67.1 \\
QV  & LongCLIP & 16 & \textbf{84.4} & \textbf{67.5} & 87.0 & 65.1 & \textbf{71.6} & \textbf{50.6} & 28.6 & \textbf{67.7} \\
QAV & CLIP     & 16 & \textbf{82.1} & 70.0 & 86.1 & \textbf{66.7} & 74.4 & 54.4 & 28.6 & 68.8 \\
QAV & LongCLIP & 16 & 81.0 & 70.0 & \textbf{86.2} & \textbf{67.6} & \textbf{75.0} & 55.2 & \textbf{29.6} & \textbf{69.2} \\
\bottomrule
\end{tabular}%
}
\end{table}

\subsection{Ablation Study for Number of Video Frames $k$} 
\label{sec:ablationStudy_of_k}
\begin{table}[H]
\centering
\caption{Question answering accuracy (\%) w.r.t. different numbers of selected frames $k$ on MLVU. LLaVA-Video-7B is used as the MLLM and LongCLIP-L is used for QAV frame selection.}
\setlength{\tabcolsep}{5pt}
\renewcommand{\arraystretch}{1.08}
\begin{tabular}{c ccccccc c}
\toprule
$k$ & TR & AR & NQA & ER & PQA & AO & AC & Avg. \\
\midrule
8  & 76.8 & 51.5 & 82.5 & 67.6 & 77.2 & 48.6 & 36.9 & 66.9 \\
16 & 79.1 & 57.0 & 84.5 & 71.0 & 80.1 & 57.1 & 48.1 & 71.3 \\
32 & \textbf{81.8} & 55.5 & 84.5 & 70.2 & 78.9 & 69.1 & 51.5 & 72.8 \\
64 & \textbf{81.8} & \textbf{59.0} & \textbf{85.1} & \textbf{73.0} & \textbf{80.5} & \textbf{74.1} & \textbf{56.3} & \textbf{75.2} \\
\bottomrule
\end{tabular}
\label{tab:mlvu_k_ablation}
\end{table}

\begin{table}[H]
\centering
\caption{Question answering accuracy (\%) w.r.t. different numbers of selected frames $k$ on Video-MME. LLaVA-Video-7B is used as the MLLM and LongCLIP-L is used for QAV frame selection.}
\setlength{\tabcolsep}{5pt}
\renewcommand{\arraystretch}{1.08}
\begin{tabular}{c cccc}
\toprule
\multirow{2}{*}{$k$} & \multicolumn{4}{c}{\textbf{Video-MME} (\textit{wo\,/\,w subs})} \\
\cmidrule(lr){2-5}
 & Overall & Short & Medium & Long \\
\midrule
8  & 59.44\,/\,62.59 & 71.11\,/\,74.44 & 58.33\,/\,60.67 & 48.89\,/\,52.67 \\
16 & 61.81\,/\,64.70 & 73.22\,/\,76.67 & 61.11\,/\,63.56 & 51.11\,/\,53.89 \\
32 & 62.74\,/\,65.81 & 75.00\,/\,77.44 & 62.22\,/\,65.89 & 51.00\,/\,54.11 \\
64 & \textbf{63.63}\,/\,\textbf{67.22} & \textbf{75.33}\,/\,\textbf{77.89} & \textbf{63.89}\,/\,\textbf{67.44} & \textbf{51.67}\,/\,\textbf{56.33} \\
\bottomrule
\end{tabular}
\label{tab:videomme_k_ablation}
\end{table}

\begin{table}[H]
\centering
\caption{Question answering accuracy (\%) w.r.t. different numbers of selected frames $k$ on LongVideoBench. LLaVA-Video-7B is used as the MLLM and LongCLIP-L is used for QAV frame selection.}
\setlength{\tabcolsep}{4pt}
\renewcommand{\arraystretch}{1.08}
\begin{tabular}{c ccccc}
\toprule
\multirow{2}{*}{$k$} & \multicolumn{5}{c}{\textbf{LongVideoBench} (\textit{wo\,/\,w subs})} \\
\cmidrule(lr){2-6}
 & Overall & 8s--15s & 15s--60s & 3m--10m & 15m--60m \\
\midrule
8  & 60.43\,/\,61.78 & \textbf{68.78}\,/\,\textbf{69.31} & \textbf{73.84}\,/\,\textbf{76.16} & 57.04\,/\,59.71 & 56.03\,/\,56.38 \\
16 & 61.33\,/\,63.87 & 67.20\,/\,69.31 & 72.67\,/\,73.26 & 59.95\,/\,64.32 & 56.91\,/\,58.87 \\
32 & \textbf{62.15}\,/\,\textbf{65.07} & 67.20\,/\,69.31 & 70.93\,/\,73.26 & \textbf{61.41}\,/\,\textbf{65.53} & \textbf{58.33}\,/\,\textbf{60.82} \\
64 & 61.11\,/\,64.25 & 67.20\,/\,69.31 & 69.19\,/\,71.51 & 59.47\,/\,62.86 & 57.80\,/\,61.35 \\\bottomrule
\end{tabular}
\label{tab:lvb_k_ablation}
\end{table}

\section{Conclusion and Future Work}
\label{conclusion}

In this paper, we presented training-free, model-agnostic frame selection methods for long-video understanding under a fixed MLLM frame budget. The first method, \textbf{QV}, ranks candidate frames by cosine similarity between frame and question embeddings. The second method, \textbf{QAV}, extends QV by scoring each frame against all question-answer pairs and taking the per-frame maximum; to the best of our knowledge, QAV is the first frame selection method to incorporate multiple-choice answer options as inference-time cues. Experiments on MLVU, Video-MME, and LongVideoBench with three open-source MLLMs show that query- and answer-aligned selection  passes more useful visual evidence to the downstream model than uniform sampling and recent training-free baselines in many settings. Future work includes extending QAV to support open-ended questions, exploring lightweight encoder adapters, and using an adaptive frame budget rather than a fixed $k$.

\bibliographystyle{plain}
\bibliography{MLLMref}

\newpage
\appendix

\end{document}